%% file: main.tex
\documentclass[keeplastbox, letterpaper, 10 pt, conference]{ieeeconf}  % Comment this line out if you need a4paper

\IEEEoverridecommandlockouts                              % This command is only needed if 
\usepackage{graphics} % for pdf, bitmapped graphics files
\usepackage{graphicx}

\usepackage{bm} %bold math (greek lowercase)
\usepackage{amsmath} % assumes amsmath package installed
\usepackage{amssymb}  % assumes amssymb package installed

\usepackage{balance}

\usepackage{float}
\usepackage{makecell}
\usepackage[normalem]{ulem}
\usepackage{hyperref}
\usepackage{color}

\usepackage{amsfonts}
\usepackage{algorithm,algpseudocode}

\usepackage{gensymb}
\usepackage[tbtags]{mathtools}

\usepackage[pdftex,dvipsnames]{xcolor}

\usepackage[draft,nomargin,inline]{fixme}
\fxsetup{theme=color}

\usepackage{todonotes}
\usepackage{units}
\usepackage{booktabs}
\usepackage{tikz}
\usepackage{textcomp}

\usepackage{import}

\usepackage[bottom]{footmisc}
\usepackage{etoolbox}

\makeatletter
\patchcmd{\@makecaption}
{\scshape}
{}
{}
{}
\makeatletter
\patchcmd{\@makecaption}
{\\}
{.\ }
{}
{}
\makeatother
\def\tablename{Table}

\usepackage[
	bibstyle=ieee, 
	citestyle=numeric-comp,
	natbib=true,
	backend=bibtex,
	firstinits,
	doi=false,
	mincitenames=1,
	maxcitenames=2,
	maxbibnames=15,
	sorting=none,
	terseinits=false,
	hyperref=true
]{biblatex}
\title{\LARGE \bf Mapless Humanoid Navigation Using Learned Latent Dynamics}

\author{Andr\'{e} Brandenburger*, Diego Rodriguez* and Sven Behnke% <-this % stops a space
\thanks{*~Authors with equal contribution. 
	All authors are with the Autonomous	Intelligent Systems (AIS) Group, Computer Science Institute VI, University of Bonn, Germany {\tt\small rodriguez@ais.uni-bonn.de}}
}%

\begin{document}

\maketitle
%\copyrightnotice
\thispagestyle{empty}
\pagestyle{empty}

%%%%%%%%%%%%%%%%%%%%%%%%%%%%%%%%%%%%%%%%%%%%%%%%%%%%%%%%%%%%%%%%%%%%%%%%%%%%%%%%
\begin{abstract}
%	Efficient and collision-free navigation is an essential requirement for deploying robots in quotidian scenarios.
%	In the robotics community, 
%	Reinforcement Learning (RL) approaches have increasingly gained popularity and have demonstrated their applicability on control tasks based on visual observations.
	In this paper, 
	we propose a novel Deep Reinforcement Learning approach to address the mapless navigation problem,
	in which the locomotion actions of a humanoid robot are taken online based on the knowledge encoded in learned models.
	Planning happens by generating open-loop trajectories in a learned latent space that captures the dynamics of the environment.
	Our planner considers visual (RGB images) and non-visual observations (e.g., attitude estimations).
	This confers the agent upon awareness not only of the scenario,
	but also of its own state.
	In addition, 
	we incorporate a termination likelihood predictor model as an auxiliary loss function of the control policy, 
	which enables the agent to anticipate terminal states of success and failure.
	In this manner,
	the sample efficiency of the approach for episodic tasks is increased.
	Our model is evaluated on the NimbRo-OP2X humanoid robot that navigates in scenes avoiding collisions efficiently in simulation and with the real hardware.
\end{abstract}

\input{introduction}
\input{related_work}
\input{background}

\input{method}

\input{experiments}

\input{conclusion}

%\flushend
\renewcommand{\bibfont}{\normalfont\footnotesize}
\balance
\printbibliography

\end{document}

%% file: introduction.tex
\section{Introduction}
\label{sec:introduction}

Mobile robot navigation typically requires a robot to traverse a series of static and dynamic obstacles in the environment to reach desired target poses,
e.g., by walking with pedestrians on sidewalks.
Traditional methods tackle this problem by processing raw sensor information (e.g., RGB images or laser scans) in order to construct local maps for path planners~\cite{Klamt2019Remote, jan2008optimal, robocup2019}.
Traditional approaches,
however,
lose expressivity with the increment of uncertainty and complexity of the environments mainly because of computational limitations associated with high-dimensional systems and real-time constraints.
In the last decade,
the rapid advances of learning methods have paved the path for an increasing development of robot learning approaches,
which are a promising alternative to solve these issues by leveraging data~\cite{alphago,alphazero,alphagozero,d4pg, worldmodels}.

In this paper, 
we address the problem of mapless navigation,
in which the robot needs to reach a known relative target pose without constructing a map of the environment.
The target pose is assumed to be given by higher-level modules (e.g., object detection, semantic segmentation or Wi-Fi signal localization).
Several Deep Reinforcement Learning (DRL) approaches have been proposed to solve this problem based on 3D data (e.g., laser scans)~\cite{zhelo2018curiosity, tai2017virtual, khan2018learning}.
In our approach,
however,
the environment is perceived by RGB-only images which, in contrast to depth data, render a harder problem for planning,
since no direct measurements to object distances are provided.
Our learned path planner considers additionally non-visual observations such as IMU measurements.
In this manner,
the planner can act upon large instabilities of the robot posture in order to avoid falls.
% IMU improves sample efficiency, falls ? -> experiment to proof this

\begin{figure}
	\centering
	\includegraphics[width=\linewidth]{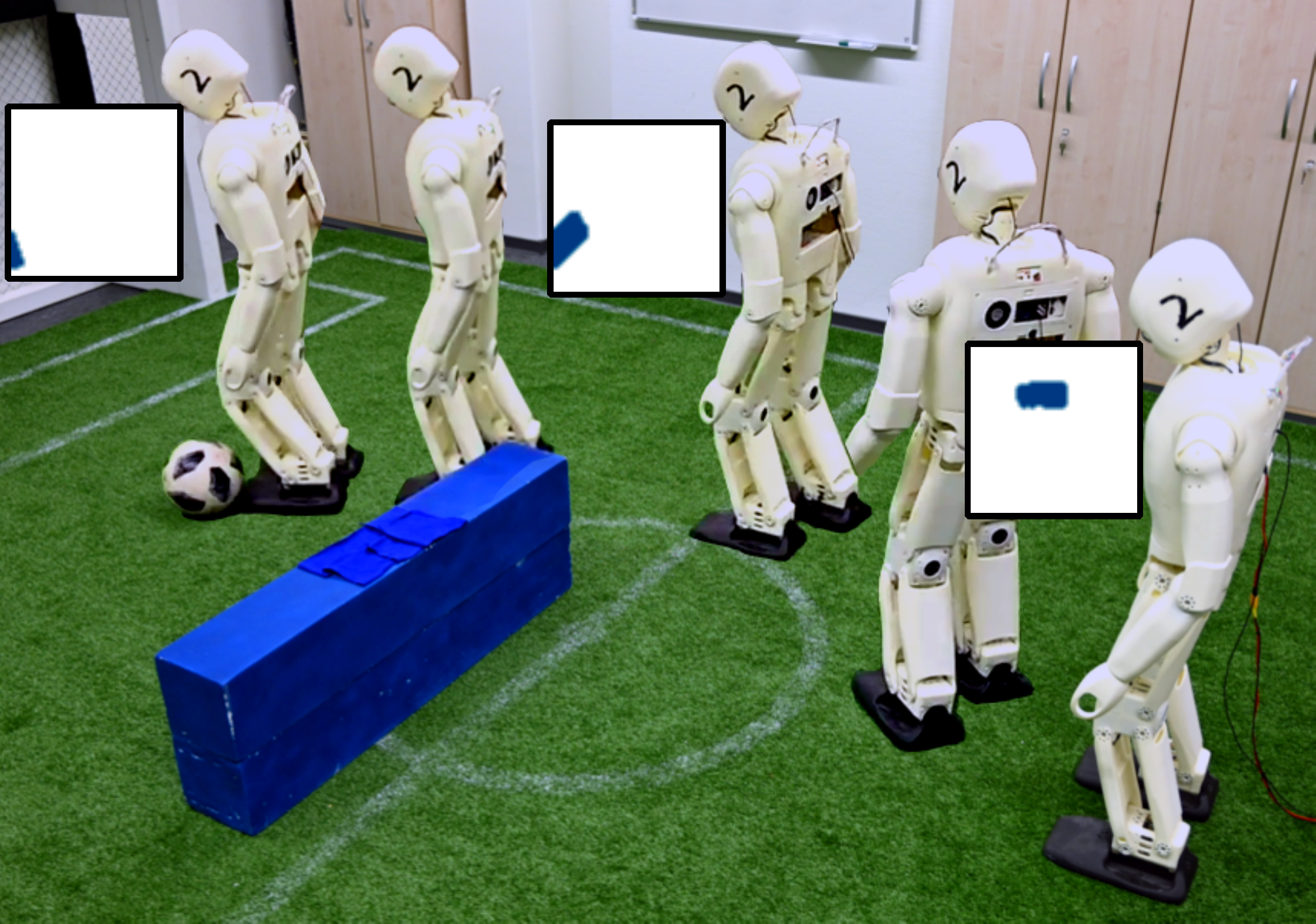}
	\caption{The NimbRo-OP2X robot navigates to reach the goal (ball) while avoiding obstacles. The actions are inferred online by a control policy (at \unit[10]{Hz}), given a segmented image (surrounded by black squares) and non-visual sensor data. 
		For clarity,
		only three segmented images are shown.
		\label{fig:teaser}
	}
	\vspace*{-3ex}
\end{figure}

Our approach is able to plan collision-free paths without local maps by learning a latent world model and by \textit{imagining} possible future outcomes based on learned models.
These open-loop (imagined) trajectories address the problem of lack of memory of Markov Decision Processes which are typically used to formulate Reinforcement Learning (RL) tasks.
Re-planning happens implicitly with each new inference step.
This allows our approach to handle uncertainty present in scenarios with dynamic obstacles.

In order to handle episodic tasks,
as the one discussed here,
we incorporate a predictor model that infers a termination likelihood and provides this information to the control policy as an auxiliary loss.
We explicitly differentiate between successful and failed terminal states;
the former encourages the agent to finish the episode collecting a high reward,
while the latter contributes to the sample efficiency and training time reduction by neglecting experiences collected during failed terminal states,
e.g.,
when the robot is lying on the floor after falling.

We evaluate our approach on a real autonomous humanoid robot (Fig.~\ref{fig:teaser}).
To handle the sim-to-real transfer,
segmented images are employed for training the learned models,
which in conjunction with noise injection and system identification allows to transfer the control policy to the real robot without retraining.

In summary, 
the main contributions of this paper are:
\begin{itemize}
	\item the formulation of a novel approach for mapless navigation that considers visual (RGB images) and non-visual observations to learn a control policy and an environment dynamics model;
	\item the introduction of a termination likelihood predictor to handle multiple terminal states specially relevant for episodic tasks;
	\item and the demonstration on a real humanoid robot of the learned policy for mapless navigation.
\end{itemize}

%% file: related_work.tex
\section{Related Work}
\label{sec:related_work}

Previous research on \textit{visual control} problems, 
in which an agent takes actions based on image observations,
has led to multiple analytical open-loop approaches~\cite{epipoles, welding, trifocal}. 
To address the typical shortcomings of analytical solutions,
especially related to the curse of dimensionality,
novel learning-based methods have called the attention of the community due to their generalization capabilities to uncertainty and due to the inference time that enables their usage in real-world tasks~\cite{depthplanning, Rodriguez2019Autonomous, devo2020towards}.
%such as poor transferability, it is possible to utilize learning based methods. The broad 
%field of Reinforcement Learning (RL) has been rapidly developing in the last 
%years, where deep neural network approaches have become more and more popular. 
%By modeling the environment as a function yielding an observation and a reward 
%at each step given an action, an agent can explore the environment with the goal 
%to optimize the expected reward. In order to work on a task, the agent can take 
%actions from the action space of the environment, influencing its state 
%evolution.

Particularly, 
RL approaches have gained increased popularity in robotics, 
where policies are learned by interaction with the environment.
Popular model-free RL methods,
such as DQN~\cite{dqn}, 
aim to construct a state-action value function (Q-value) that quantifies the quality of state-action pairs to maximize an accumulative reward in the long term~\cite{alphago, alphagozero, alphazero}.
Other model-free RL approaches,
called policy gradient methods,
construct a policy by optimizing a cost function directly,
such as D4PG~\cite{d4pg} and PPO~\cite{schulman2017proximal}.
Although these RL methods have been successfully implemented in robotics applications~\cite{khan2018learning, rodriguez2020} including visual control tasks~\cite{depthplanning,naos}, 
the training with raw images requires a large amount of data---due to the absence of a learned dynamics model, 
which could encode the state evolution effectively.

While model-free RL approaches are often straightforward to employ,
model-based methods can be more sample-efficient by exploiting a learned dynamics model.
One of the first attempts to learn a control policy in conjunction with a dynamics model is Dyna-Q~\cite{sutton2018reinforcement}.
Recent approaches such as~\cite{worldmodels},~\cite{planet} and~\cite{dreamer} are able to process raw image observations directly by using self-supervised representation techniques, i.e., autoencoders.
Inspired by these works,
in this paper,
we present a novel model-based RL approach for mapless navigation.

Mapless navigation using RL have been previously addressed~\cite{tai2017virtual,khan2018learning,zhelo2018curiosity}. 
\citet{khan2018learning} proposed a two-stage architecture consisting of local planners defined by value iteration networks and differentiable memory networks that provide past information.
~\citet{zhelo2018curiosity} do not define any memory component but encourage curiosity-based exploration formulated in a secondary reward function, 
and consequently the agent is able to navigate in long corridors and dead corners.
None of these approaches, 
however,
are able to handle dynamic obstacles and require depth data as input.
Moreover, 
these approaches were evaluated in known scenarios only,
thus their generalization capabilities are questionable.

Few RL approaches for robot navigation based on RGB images have been demonstrated in real robots~\cite{depthplanning,naos}.
\citet{depthplanning} propose a depth prediction network based on monocular RGB images that infers a depth field and a Q-value function for controlling a mobile robot.
\citet{naos} investigate visual navigation on a bipedal platform and learned a control policy by using DDPG.
None of these approaches, 
however, 
incorporate latent dynamics models and terminal states for episodic tasks are not explicitly handled.
%mainly because of the more stable gait of the small NAO robots.

%% file: background.tex
\section{Background}
\label{sec:background}

%In this section, we will discuss the fundamental architecture of our approach, 
%which is based on the popular \textit{Dreamer} model\cite{dreamer}.
As common in RL, 
we model the environment as a Markov 
Decision Process (MDP) described by a tuple $(S, A, P, R, \gamma)$ of environment states 
$S$,
action space $A$, 
state transition probabilities ${P:S\times A\times S\to[0,1]}$,
reward function ${R:S\times A\to\mathbb{R}}$, 
and discounted factor $\gamma\in[0,1]$. 
The goal of the agent is to take actions $a_t\in A$ that maximize the expected reward. 
%While the state transitions $P$ and the reward function $R$ depend on the true 
%environment state, 
Often, 
the agent only has access to partial observations $o_t\in O$ of the environment, 
which are provided according to state observation probabilities ${\Omega:S\times O\to [0,1]}$. This results in a 
Partially Observable Markov Decision Process (POMDP) defined by $(S, 
A, P, R, \gamma, O, \Omega)$.

%Model-free methods directly define policies ${\Pi_F:O\to A}$ on the 
%observation space $O$, 
%while model-based approaches make use of a learned environment model.
%This model constructs an environment representation $W$ that aims to resemble the 
%underlying environment state. Consequently, the policy $\Pi:W\to A$ is defined 
%on the inferred state of the world model.

In domains % no comma is fine here
where the observations are defined as images,
policies are often expensive to train due to the high dimensionality of the observation space $O$. 
Thus, 
representation techniques such as autoencoders (${\mathcal{E}:O\to W}$) are frequently incorporated to reduce the dimensionality of the image input and to define a latent state $W$ of the environment model~\cite{worldmodels,planet,dreamer}. 
The latent state dynamics ${\mathcal{D}:W\times A\to W}$ can be learned effectively to resemble the unknown true state transition $P$ of the environment. 
Both, 
the autoencoder and the latent state dynamics can be combined to form a non-linear Kalman Filter,
in which the state prediction $\tilde{w}$ is given by $\mathcal{D}$,
and the filtering is done by the encoder $\mathcal{E}$~\cite{planet}.

\begin{figure*}
	\centering
	\def\svgwidth{\linewidth}
	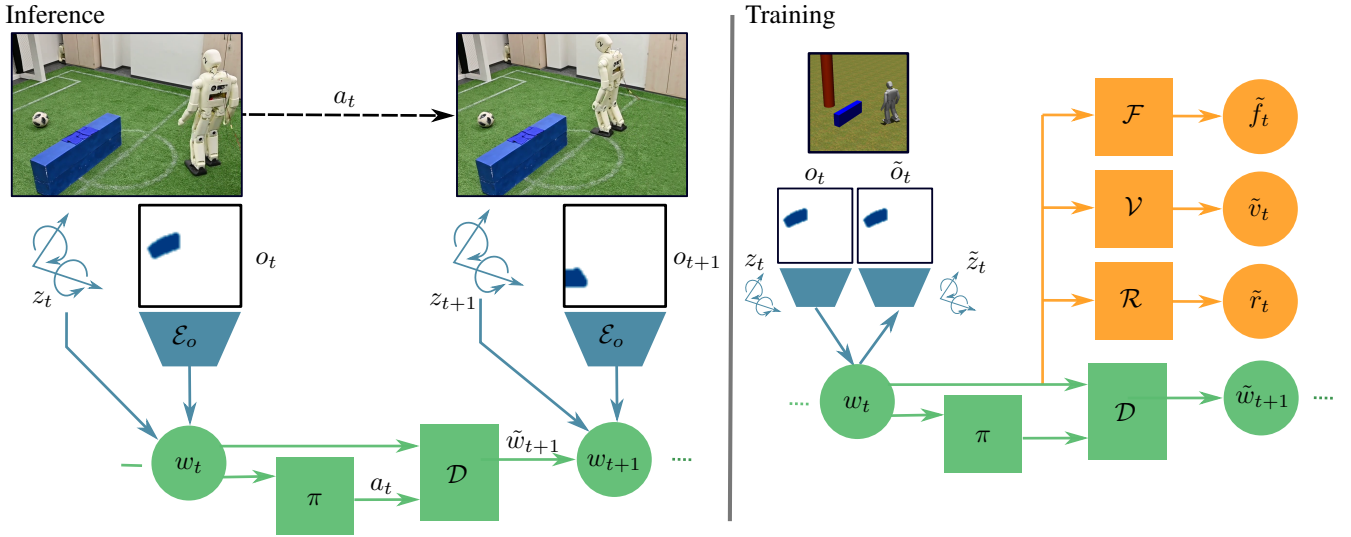
	\caption{
		Approach overview. 
		Visual $o_t$ and non-visual $z_t$ observations are fused into a latent vector $w_t$ which is used by the policy $\pi$ to infer actions $a_t$.
		The dimensionality of the images is reduced by employing an autoencoder $\mathcal{E}_o$.
		The dynamics model $\mathcal{D}$ predicts the next latent state $\tilde{w}_{t+1}$ which is later filtered by sensor data observed in $t+1$,
		namely by $z_{t+1}$ and by $\mathcal{E}_o(o_{t+1})$.
		During training, 
		a decoder is also learned which aims to reconstruct an observed image $o_t$ from $w_t$.
		Additionally, 
		the state value $\mathcal{V}$, the reward $\mathcal{R}$, and the termination likelihood $\mathcal{F}$ predictors are learned, 
		which are used in the loss function of the policy $\mathcal{\pi}$ (Eq.~(\ref{eq:actor_loss})).
	}
	\label{fig:model}
	\vspace*{-2ex}
\end{figure*}

\citet{dreamer} recently proposed a model-based RL approach that builds a latent space $W$ and dynamics model $\mathcal{D}$ which are ultimately employed in an open-loop fashion to plan latent trajectories $\tilde{w}_{t_0}\dots \tilde{w}_{t_N}$.
%We will briefly introduce the Dreamer model since it is the backbone of our approach.
For each of the latent states $w_{t_i}$, 
a state value is calculated by use of the Bellman return:
\begin{equation}
\label{eq:bellman}
	v_{t_i} = \sum_{t = t_i}^{t_N} \gamma^{(t-t_i)} \tilde{r}_t\,,
\end{equation}
given predicted rewards $\tilde{r}_t$ inferred by a predictor ${\mathcal{R}:W\to\mathbb{R}}$.
Additionally,
a value predictor model ${\mathcal{V}:W\to\mathbb{R}}$ is incorporated to optimize the Bellman consistency. 
%To address approximation errors and support generality
The predicted rewards $\tilde{r}_t$,
values $\tilde{v}_t$ and actions $a_t$ are modeled stochastically.
More precisely, 
the means of Gaussian distributions are dictated by the prediction models $\mathcal{R}, \mathcal{V}, \pi$. 
The standard deviations of the action distributions are also inferred by the actor $\pi$, 
while a unit standard deviation is chosen for the other predictors.
%Using this stochastic representation, the prediction models are 
%trained by the negative log likelihood of the true data. 
In contrast to the fully stochastic prediction models, 
the latent state is constructed using the Recurrent State Space Model (RSSM), 
which represents the latent space by a mixture of deterministic and stochastic states~\cite{planet, doerr2018}. 

The autoencoder $\mathcal{E}$ as well as the prediction model $\mathcal{R}$ are trained using the negative log likelihood of the true data from an experience replay buffer.
In addition, 
the latent state dynamics loss is based on the Kullback-Leibler divergence between the open-loop 1-step prediction and the closed-loop 1-step prediction. 
% This ensures that the dynamics model $\mathcal{D}$ resembles the Kalman filtering process. % This does not give a lot of information to the reader IMO
The loss is based on the Information Bottleneck objective~\cite{ibottleneck},
defined as:
\begin{equation}
\label{eq:latent_loss}
\begin{split}
 &\mathcal{L}_{\mathcal{E}, \mathcal{R}, \mathcal{D}}
= 
-\mathbb{E}
\Big[ 
\sum_t
      \ln \mathcal{E}\left( o_t \vert w_t \right) 
      + \ln \mathcal{R}\left( r_t \vert w_t \right)
\\
      &- \beta \, \text{KL}\left[
           \mathcal{E}\left( w_t \vert 
            \mathcal{D}\left(\tilde{w}_{t}\vert w_{t-1}, a_{t-1}\right), o_t 
        \right)
          \vert \vert
          \mathcal{D}\left( \tilde{w}_t \vert w_{t-1}, a_{t-1} \right)
        \right]
\Big] \,.
%		&- \beta \, \text{KL}\left[
%		\mathcal{E}\left( \tilde{w}_{t}\vert w_{t-1}, a_{t-1}, o_t \right)
%		\vert \vert
%		\mathcal{D}\left( \tilde{w}_t \vert w_{t-1}, a_{t-1} \right)
%		\right]
%		\Big] \,.
\end{split}
\end{equation}

In contrast to the $\mathcal{E}, \mathcal{R}$ and $\mathcal{D}$ networks, 
the value model $\mathcal{V}$ and the actor $\pi$ are not trained on the recorded episodes, 
but on state trajectories that are generated through consecutive inference of the learned latent state dynamics $\mathcal{D}$ in conjunction with the actor $\pi$ on a single filtered state posterior. 
This results in a tuple  $(w_{t_i},\dots, w_{t_{i+H}})$ of unfiltered states that are used to train the value and actor models over a horizon of length $H$. 
The loss of the value prediction network minimizes the regression error of the state value that is calculated via reward predictions:
\begin{equation}
\label{eq:v_loss}
 \mathcal{L}_\mathcal{V} = -\mathbb{E}\left[ \sum_{t=t_i}^{t_{i+H}}\left\Vert\mathcal{V}(w_t)- v_t\right\Vert^2 \right],
\end{equation}

while the loss of the actor $\pi$ maximizes the value of the generated state tuples:
\begin{equation}
 \mathcal{L}_\pi = \mathbb{E}\left[ \sum_{t=t_i}^{t_{i+H}}v_t \right].
\end{equation}

%% file: figs/compiled/model_3.pdf_tex
%% Creator: Inkscape inkscape 0.92.3, www.inkscape.org
%% PDF/EPS/PS + LaTeX output extension by Johan Engelen, 2010
%% Accompanies image file 'model_3.pdf' (pdf, eps, ps)
%%
%% To include the image in your LaTeX document, write
%%   \input{<filename>.pdf_tex}
%%  instead of
%%   \includegraphics{<filename>.pdf}
%% To scale the image, write
%%   \def\svgwidth{<desired width>}
%%   \input{<filename>.pdf_tex}
%%  instead of
%%   \includegraphics[width=<desired width>]{<filename>.pdf}
%%
%% Images with a different path to the parent latex file can
%% be accessed with the `import' package (which may need to be
%% installed) using
%%   \usepackage{import}
%% in the preamble, and then including the image with
%%   \import{<path to file>}{<filename>.pdf_tex}
%% Alternatively, one can specify
%%   \graphicspath{{<path to file>/}}
%% 
%% For more information, please see info/svg-inkscape on CTAN:
%%   http://tug.ctan.org/tex-archive/info/svg-inkscape
%%
\begingroup%
  \makeatletter%
  \providecommand\color[2][]{%
    \errmessage{(Inkscape) Color is used for the text in Inkscape, but the package 'color.sty' is not loaded}%
    \renewcommand\color[2][]{}%
  }%
  \providecommand\transparent[1]{%
    \errmessage{(Inkscape) Transparency is used (non-zero) for the text in Inkscape, but the package 'transparent.sty' is not loaded}%
    \renewcommand\transparent[1]{}%
  }%
  \providecommand\rotatebox[2]{#2}%
  \newcommand*\fsize{\dimexpr\f@size pt\relax}%
  \newcommand*\lineheight[1]{\fontsize{\fsize}{#1\fsize}\selectfont}%
  \ifx\svgwidth\undefined%
    \setlength{\unitlength}{1120.63402318bp}%
    \ifx\svgscale\undefined%
      \relax%
    \else%
      \setlength{\unitlength}{\unitlength * \real{\svgscale}}%
    \fi%
  \else%
    \setlength{\unitlength}{\svgwidth}%
  \fi%
  \global\let\svgwidth\undefined%
  \global\let\svgscale\undefined%
  \makeatother%
  \begin{picture}(1,0.39666704)%
    \lineheight{1}%
    \setlength\tabcolsep{0pt}%
    \put(0,0){\includegraphics[width=\unitlength,page=1]{model_3.pdf}}%
    \put(0.14329417,0.04905766){\color[rgb]{0,0,0}\makebox(0,0)[t]{\lineheight{1.25}\smash{\begin{tabular}[t]{c}$w_t$\end{tabular}}}}%
    \put(0,0){\includegraphics[width=\unitlength,page=2]{model_3.pdf}}%
    \put(0.23735563,0.02229753){\color[rgb]{0,0,0}\makebox(0,0)[t]{\lineheight{1.25}\smash{\begin{tabular}[t]{c}$\pi$\end{tabular}}}}%
    \put(0.3437879,0.03903889){\color[rgb]{0,0,0}\makebox(0,0)[t]{\lineheight{1.25}\smash{\begin{tabular}[t]{c}$\mathcal{D}$\end{tabular}}}}%
    \put(0,0){\includegraphics[width=\unitlength,page=3]{model_3.pdf}}%
    \put(0.46131694,0.05173468){\color[rgb]{0,0,0}\makebox(0,0)[t]{\lineheight{1.25}\smash{\begin{tabular}[t]{c}$w_{t+1}$\end{tabular}}}}%
    \put(0,0){\includegraphics[width=\unitlength,page=4]{model_3.pdf}}%
    \put(0.28742236,0.03382681){\color[rgb]{0,0,0}\makebox(0,0)[t]{\lineheight{1.25}\smash{\begin{tabular}[t]{c}$a_t$\end{tabular}}}}%
    \put(0,0){\includegraphics[width=\unitlength,page=5]{model_3.pdf}}%
    \put(0.5236087,0.20288418){\color[rgb]{0,0,0}\makebox(0,0)[t]{\lineheight{1.25}\smash{\begin{tabular}[t]{c}$o_{t+1}$\end{tabular}}}}%
    \put(0,0){\includegraphics[width=\unitlength,page=6]{model_3.pdf}}%
    \put(0.34100481,0.17209768){\color[rgb]{0,0,0}\makebox(0,0)[t]{\lineheight{1.25}\smash{\begin{tabular}[t]{c}$z_{t+1}$\end{tabular}}}}%
    \put(0.45941873,0.14343031){\color[rgb]{0,0,0}\makebox(0,0)[t]{\lineheight{1.25}\smash{\begin{tabular}[t]{c}$\mathcal{E}_o$\end{tabular}}}}%
    \put(0,0){\includegraphics[width=\unitlength,page=7]{model_3.pdf}}%
    \put(0.26003003,0.32319642){\color[rgb]{0,0,0}\makebox(0,0)[t]{\lineheight{1.25}\smash{\begin{tabular}[t]{c}$a_t$\end{tabular}}}}%
    \put(0,0){\includegraphics[width=\unitlength,page=8]{model_3.pdf}}%
    \put(0.0433471,0.38271619){\color[rgb]{0,0,0}\makebox(0,0)[t]{\lineheight{1.25}\smash{\begin{tabular}[t]{c}Inference\end{tabular}}}}%
    \put(0.59315729,0.38310747){\color[rgb]{0,0,0}\makebox(0,0)[t]{\lineheight{1.25}\smash{\begin{tabular}[t]{c}Training\end{tabular}}}}%
    \put(0,0){\includegraphics[width=\unitlength,page=9]{model_3.pdf}}%
    \put(0.40086839,0.06518678){\color[rgb]{0,0,0}\makebox(0,0)[t]{\lineheight{1.25}\smash{\begin{tabular}[t]{c}$\tilde{w}_{t+1}$\end{tabular}}}}%
    \put(0.2013573,0.20288414){\color[rgb]{0,0,0}\makebox(0,0)[t]{\lineheight{1.25}\smash{\begin{tabular}[t]{c}$o_{t}$\end{tabular}}}}%
    \put(0,0){\includegraphics[width=\unitlength,page=10]{model_3.pdf}}%
    \put(0.03389892,0.17343623){\color[rgb]{0,0,0}\makebox(0,0)[t]{\lineheight{1.25}\smash{\begin{tabular}[t]{c}$z_{t}$\end{tabular}}}}%
    \put(0.14043233,0.14319371){\color[rgb]{0,0,0}\makebox(0,0)[t]{\lineheight{1.25}\smash{\begin{tabular}[t]{c}$\mathcal{E}_o$\end{tabular}}}}%
    \put(0,0){\includegraphics[width=\unitlength,page=11]{model_3.pdf}}%
    \put(0.61174675,0.26565612){\color[rgb]{0,0,0}\makebox(0,0)[t]{\lineheight{1.25}\smash{\begin{tabular}[t]{c}$o_t$\end{tabular}}}}%
    \put(0.84916953,0.3079288){\color[rgb]{0,0,0}\makebox(0,0)[t]{\lineheight{1.25}\smash{\begin{tabular}[t]{c}$\mathcal{F}$\end{tabular}}}}%
    \put(0.84964234,0.23930891){\color[rgb]{0,0,0}\makebox(0,0)[t]{\lineheight{1.25}\smash{\begin{tabular}[t]{c}$\mathcal{V}$\end{tabular}}}}%
    \put(0.84774885,0.17021634){\color[rgb]{0,0,0}\makebox(0,0)[t]{\lineheight{1.25}\smash{\begin{tabular}[t]{c}$\mathcal{R}$\end{tabular}}}}%
    \put(0.94366928,0.30864957){\color[rgb]{0,0,0}\makebox(0,0)[t]{\lineheight{1.25}\smash{\begin{tabular}[t]{c}$\tilde{f}_{t}$\end{tabular}}}}%
    \put(0.94366882,0.23861007){\color[rgb]{0,0,0}\makebox(0,0)[t]{\lineheight{1.25}\smash{\begin{tabular}[t]{c}$\tilde{v}_{t}$\end{tabular}}}}%
    \put(0.94414148,0.16951734){\color[rgb]{0,0,0}\makebox(0,0)[t]{\lineheight{1.25}\smash{\begin{tabular}[t]{c}$\tilde{r}_{t}$\end{tabular}}}}%
    \put(0,0){\includegraphics[width=\unitlength,page=12]{model_3.pdf}}%
    \put(0.67595278,0.26658124){\color[rgb]{0,0,0}\makebox(0,0)[t]{\lineheight{1.25}\smash{\begin{tabular}[t]{c}$\tilde{o}_t$\end{tabular}}}}%
    \put(0,0){\includegraphics[width=\unitlength,page=13]{model_3.pdf}}%
    \put(0.64304999,0.09508683){\color[rgb]{0,0,0}\makebox(0,0)[t]{\lineheight{1.25}\smash{\begin{tabular}[t]{c}$w_t$\end{tabular}}}}%
    \put(0,0){\includegraphics[width=\unitlength,page=14]{model_3.pdf}}%
    \put(0.73711179,0.06966521){\color[rgb]{0,0,0}\makebox(0,0)[t]{\lineheight{1.25}\smash{\begin{tabular}[t]{c}$\pi$\end{tabular}}}}%
    \put(0.84354408,0.08506805){\color[rgb]{0,0,0}\makebox(0,0)[t]{\lineheight{1.25}\smash{\begin{tabular}[t]{c}$\mathcal{D}$\end{tabular}}}}%
    \put(0,0){\includegraphics[width=\unitlength,page=15]{model_3.pdf}}%
    \put(0.94636197,0.09871507){\color[rgb]{0,0,0}\makebox(0,0)[t]{\lineheight{1.25}\smash{\begin{tabular}[t]{c}$\tilde{w}_{t+1}$\end{tabular}}}}%
    \put(0,0){\includegraphics[width=\unitlength,page=16]{model_3.pdf}}%
    \put(0.73229772,0.198782){\color[rgb]{0,0,0}\makebox(0,0)[t]{\lineheight{1.25}\smash{\begin{tabular}[t]{c}$\tilde{z}_t$\end{tabular}}}}%
    \put(0.56644464,0.20130475){\color[rgb]{0,0,0}\makebox(0,0)[t]{\lineheight{1.25}\smash{\begin{tabular}[t]{c}$z_t$\end{tabular}}}}%
    \put(0,0){\includegraphics[width=\unitlength,page=17]{model_3.pdf}}%
  \end{picture}%
\endgroup%

%% file: method.tex
\section{Method}
\label{sec:method}
Our approach aims to solve the mapless navigation problem.
We fuse visual $o_t$ and non-visual observations $z_t$ into a latent state $w_t$.
Actions $a_t$ are taken by a learned policy $\pi$ that is trained by open-loop (imagined) latent trajectories inferred by an environment dynamics model $\mathcal{D}$.
State value $\mathcal{V}$, 
reward $\mathcal{R}$,
and termination likelihood $\mathcal{F}$ predictors are also learned during training.
The latter is employed to increase the sample efficiency of this episodic task.
The high dimensionality of images is reduced by the incorporation of a variational autoencoder $\mathcal{E}_o$.
Fig.~\ref{fig:model} shows an overview of our approach;
the algorithm is summarized in Algorithm.~\ref{alg:approach}.

%We use Dreamer~\cite{dreamer}, 
%briefly explained in Section~\ref{sec:background},
%as the backbone of our approach.
%In contrast to Dreamer, our method
%is able to handle non-visual observations and episodic tasks through the introduction of a termination likelihood predictor.
%Additionally,
%the learned policy is successfully transferred to a real humanoid robot.

\subsection{Observation Model}
%In contrast to previous works, 
We enrich the observation space by considering both, image inputs and non-visual sensory data. % for me its easier to read with comma, otherwise I read it like 'both image inputs'
Consequently,
we propose two separate autoencoders,
$\mathcal{E}_o$ and $\mathcal{E}_z$, 
where $\mathcal{E}_o$ represents the convolutional autoencoder for image inputs and $\mathcal{E}_z$ processes non-visual sensor information. 
Since $z_t$ is low dimensional, we forego the encoder part of $\mathcal{E}_z$ and only utilize the decoder to predict the measurements from the latent state.
The autoencoders $\mathcal{E}_o$ and $\mathcal{E}_z$ are trained by optimizing the negative log likelihood of the true observations under the observation models:
\begin{equation}
	\mathcal{L}_{\mathcal{E}_z, \mathcal{E}_o} = - \mathbb{E} \left[ \sum_t \ln \mathcal{E}_o(o_t \vert w_t) + \ln \mathcal{E}_z(z_t \vert w_t) \right] .
\end{equation}

\input{figs/compiled/algorithm}

\subsection{Termination Likelihood Predictor}
A terminal state can represent either success or failure in episodic tasks.
Typical such tasks define a successful terminal state that indicates the achievement of the task's goal.
Moreover,
early termination is an established strategy for improving sample efficiency,
so that an episode is terminated when certain states are reached whose contribution is considered negligible,
e.g., states that represent a biped robot lying on the floor in a navigation task.
In this manner,
the sample acquisition time and the corresponding gradient propagation are avoided for these terminal states that do not contribute to reaching the task's goal.

Having multiple terminal states $f_i$ poses a challenge to the design of the reward function, 
as it is no longer possible to reward or to penalize termination per se.
Due to the different nature of each terminal state, 
success and failure need to be addressed separately.
In episodic tasks, 
one issue with successful termination emerges when the agent prefers to collect rewards instead of terminating because it continues accumulating reward.
While termination rewards at the end of an episode promise a fast and straightforward solution to this issue, 
their inherent discontinuity makes them hard to predict.
Thus, 
we introduce a termination likelihood model, 
which predicts a continuous indicator $f_{i,t}$ for reaching a terminal state.
In contrast to $\mathcal{R},\mathcal{V}$ and $\pi$, 
the termination likelihood is modeled as beta distributed.
The inferred termination likelihood is weighed and passed as a smooth learning signal to the actor model,
enabling the agent to anticipate success and failure states. 
The actor loss is reformulated as:
\begin{equation}
	\label{eq:actor_loss}
	\mathcal{L}_\pi = - \mathbb{E} \left[\sum_t \left( v_t + \sum_i \lambda_i \mathcal{F}(f_{i,t}\vert w_t) \right) \right] \,.
\end{equation}

While the Dreamer~\cite{dreamer} also computes an approximation of the termination probability, 
it is aimed to weigh down the return, as the agent cannot collect rewards after termination. 
However, 
this reward weighing reduces the effect of constant termination rewards when reaching the goal.
In contrast, 
we let the termination likelihood influence the actor loss directly, 
without influencing the per-step-reward.
Furthermore, 
we consider the semantically different failure and success termination states separately.

\subsection{Task Definition}
%% Mail goal
The goal of the agent is to reach a desired 2D pose on a flat ground plane without collisions with obstacles in the environment.
%% State space
The agent perceives the environment through RGB images and additional non-visual sensors.
The images are taken from an ego perspective of a walking humanoid robot and,
hence,
contain much walking-induced motion.
Similar to \cite{naos}, they are passed through a semantic segmentation module that classifies obstacles pixelwise.
Unnecessary textural information and background pixels are therefore removed.
This image segmentation facilitates the image prediction and the real-world transfer.
The resulting segmented image (resolution ${64\times64}$ in our experiments) defines the visual observations $o_t$ of our approach.

In addition,
the non-visual observation is defined as ${z_t = [\mathbf{V}_t, h_t, d_t, \theta_t, \mathbf{R}_t]^T}$,
where $\mathbf{V}_t$ is the current gait velocity,
$h_t$ is the yaw joint position of the head, 
$[d_t, \theta_t]^T$ is the relative target position expressed in polar coordinates,
and $\mathbf{R}_t$ contains the pitch and roll rotation of the robot base link.
Note that in real world applications,
the relative target position is often determined by high level-task planners or perception modules.

%% Action space
In each time step, 
the agent selects an action $a_t=[\Delta \mathbf{V}_t,\Delta h_t]^T$,
where $\Delta \mathbf{V}_t$ is an increment of the gait velocity, 
i.e., 
${\mathbf{V}_{t+1} = \mathbf{V}_t + \Delta \mathbf{V}_t}$,
and $\Delta h_t$ represents an increment of the yaw head position.
Note that the incremental action representation is introduced to guide the agent learning process, 
especially at the beginning of training where exploration of the action space might lead to oscillating motions that saturate the low-level joint controllers.
The velocity vector $\mathbf{V}_t=[v_x,v_y,\omega_z]^T$ consists of the translational $x$- and $y$-velocities,
as well as a rotational velocity around the $z$-axis of the robot.
Overall,
a 4D action space is defined.

\subsection{Terminal States}
We propose two different termination criteria. 
The robot arrives into a successful terminal state when the distance $d_t$ to the target is below a certain threshold,
whereas the failure terminal state is reached when the sum of the absolute roll and 
pitch rotations $\vert \mathbf{R}_{0,t} \vert + \vert \mathbf{R}_{1,t} \vert$ of 
the robot surpasses limit values that indicate an imminent robot fall. 
Both error values,
i.e., 
the distance and orientation errors,
are passed through an exponential decay to yield a continuous signal ${f_{i,t}}$ that indicates termination whenever ${f_{i,t} = 1}$. 
Note that the causality ${f_{i,t} = 1 \implies \forall \Delta t > 0: f_{i,t+\Delta t} = 1}$ holds, 
which can be incorporated into the latent world model.

\subsection{Reward Function}
We define the task reward at time $t$,
${r_t=\sum_{i=0}^N \eta_i r_{i,t}}$ as the weighted sum of $N$ sub-reward terms.
For brevity,
the dependence of time will be dropped in the equations.

The main sub-rewards encourage the agent to reach the target pose and are formulated as:
\begin{align}
  r_{d} = 1-\frac{d}{d_0} &\hspace{1em} \in (-\infty, 1],\\
  r_{\theta} = -\left| \frac{\theta}{\pi} \right| &\hspace{1em} \in [-1, 0]\,,
\end{align}
where $d$ is the distance to the target position,
$d_0$ is the distance from the initial pose to the target,
and $\theta\in(-\pi,\pi]$ is the relative orientation of the robot to the target position,
for example, 
${\theta=0}$ means the robot is directly facing the target position.
The former sub-reward encourages the agent to walk towards the target by reducing 
the distance $d$, 
while the latter penalizes the robot when it is not facing the target.

In addition, 
we define a sub-reward based on the location of the target inside the robot's camera image. 
This {target attention} reward is generated by the multiplication of an importance map $\mathbf{L} \in [0,1]^{64 \times 64}$ 
with a binary segmented image $\mathbf{I} \in \{0, 1\}^{64 \times 64}$ showing 
only the target:
\begin{equation}
 r_{a} = \frac{\sum_{i,j} \mathbf{L}_{i,j} \mathbf{I}_{i,j}}{\sum_{i,j} \mathbf{I}_{i,j}} \in [0, 1]\,.
\end{equation}
%By normalizing using the total number of active pixels in $\mathbf{I}$, we 
%conclude with the average activated importance as the sub-reward. 
We set $r_{a} = 0$ if the target is not visible in the image, i.e., $\sum_{i,j} 
\mathbf{I}_{i,j} = 0$. To ensure that the agent prefers to keep the target in 
the center of the observed egocentric images, we set ${\mathbf{L}_{i,j} = 1}$ 
for pixels at the center while we quadratically discount the values towards 
${\mathbf{L}_{i,j} = 0}$ at the borders of the image. 
Consequently, 
the agent will try to keep the target in the center of its field of view mainly by controlling the head yaw joint, 
whose motion relates directly with the relative movement of the target in the observed images. 
Note that this target relative position is also affected by the gait. 
%Encouraging the policy to keep the target in the center of its field of view minimize the chance to lost the target during movement,
%which would necessitate the expensive relocation of the goal. 

In order to avoid oscillating motions of the head, 
we penalize the normalized head position $h$ and the normalized head control action $\Delta h$ quadratically:
\begin{align}
 r_{h} = -(\Delta h)^2 &\in [-1, 0],\\
 r_{H} = -(h)^2 &\in [-1, 0]\,.
\end{align}

In addition, we encourage the agent to maintain a safe distance to obstacles by 
penalizing its distance towards the closest obstacle $\rho$:
\begin{equation}
 r_\rho = \text{clip}\left(-(1-\rho), -1, 0\right) \in [-1, 0]\,.
\end{equation}

Finally, 
we penalize the current gait velocity using a sigmoid kernel $k(x) = 1/[1+\exp(-\alpha x -c)]$ to limit the maximum gait velocity of the agent due to difference between the simulated and the real gait.
This penalization is formulated as:
\begin{equation}
 r_{v} = 1-k(\vert\vert \mathbf{V}\vert\vert_2)\,.
\end{equation}

%% file: figs/compiled/algorithm.tex
 \begin{algorithm}[]
\caption{Training}
\label{alg:approach}
\begin{algorithmic}[1]

\State Initialize $\mathcal{E}, \mathcal{D}, \mathcal{F}, \mathcal{V}, \mathcal{R}, \pi$
\State $E \leftarrow \{\}$
\vspace{2mm}
%\State Generate environment scenes
\State $e_0 \leftarrow$ Collect initial episodes using a random agent
\State $E \leftarrow E \cup e_0$
\vspace{2mm}
\While{training}
  \State Draw random subset $\hat{E}$ from $E$
  \State Fit $\mathcal{E}, \mathcal{D}, \mathcal{R}, \mathcal{F}$ on $\hat{E}$ (similar to Eq.~\eqref{eq:latent_loss})
  \vspace{2mm}
  \State $\hat{E}^+ \leftarrow$ Extrapolate each state in $\hat{E}$ using $\pi$ and $\mathcal{D}$
  \State $\tilde{r}^+ \leftarrow$ Predict rewards in $\hat{E}^+$ using $\mathcal{R}$
  \State $v^+ \leftarrow$ Calculate returns from $\tilde{r}^+$ (see Eq.~\eqref{eq:bellman})
  \State $\tilde{v}^+ \leftarrow$ Predict value using $\mathcal{V}$
  \vspace{2mm}
  \State Fit $\mathcal{V}$ on $v^+$ (see loss in Eq.~\eqref{eq:v_loss})
  \State Maximize $\pi$ w.t. $\tilde{v}^+$ (see loss in Eq.~\eqref{eq:actor_loss})
  \vspace{2mm}
%  \State Generate environment scenes
  \State $e \leftarrow$ Generate episodes from policy
  \State $E \leftarrow E \cup e$
  
\EndWhile
\end{algorithmic}
\end{algorithm}

%% file: experiments.tex
\section{Evaluation}
\label{sec:evaluation}

We evaluate our approach on the NimbRo-OP2X humanoid robot~\cite{op2x}.
All training is done using experience collected only in simulation employing MuJoCo as multi-body simulator.
Eight environments are executed in parallel to speed up the data acquisition.
The policy frequency is \unit[10]{Hz}, 
while the simulation runs at \unit[1]{kHz}.
The robot incorporates a bipedal gait engine, 
which generates leg motions based on a target gait velocity $\mathbf{V}_t$~\cite{op2x, philipp}.
The gait runs at \unit[100]{Hz}.
For collision checking operations, 
the robot links are approximated by geometrical primitives.

Each episode starts with the humanoid robot standing without any obstacle in its direct vicinity.
The target position, 
the number of obstacles, 
their poses and geometries are drawn uniformly at random.
To encourage the development of robot skills to circumvent obstacles,
each episode places an obstacle between the initial position of the robot and the target pose with probability $p_{block}=0.5$.
The feasibility of reaching the target is checked by an A* planner;
if no path is found,
a new environment is generated.
The agent has a maximum time of \unit[60]{s} to complete the task before the episode ends.

%The agent captures a monocular color image ($64\times64$).
In simulation, 
the renderer provides the segmented image ($64\times64$), 
while a threshold in the HSV space is used for the real world experiments. 
For applications with non-distinctive object colors, segmentation approaches, such as \cite{Long_2015_CVPR}, could be applied.
The inferred actions are bounded to $\Delta \mathbf{V}\in[-0.06,0.06]^3$ and $\Delta h\in[-0.012,0.012]$.
The weights of the reward functions are: 
${\eta_d=1.0}$, ${\eta_\theta=0.2}$, ${\eta_a=0.2}$, ${\eta_{h}=0.08}$, ${\eta_{H}=0.08}$, $\eta_\rho = 0.2$, and ${\eta_v=0.35}$, 
and the weights of the terminal states are:
${\lambda_s=-1150}$ and ${\lambda_f=3250}$.
The models are trained every 4,000 recorded steps.
Other hyperparameters match the Dreamer model~\cite{dreamer}.
The policy is trained for 2 million simulation steps,
resulting in a total training time of around 2 days on a computer with an Intel i9-9990K CPU, 
64GB of RAM and an nVidia GeForce 2080 Ti with 12GB of VRAM.
This model is able to solve simple scenes after 200.000 steps, 
i.e., 
less than five hours of training.

\begin{figure}[b!]
	\centering
	\includegraphics[width=1.0\linewidth]{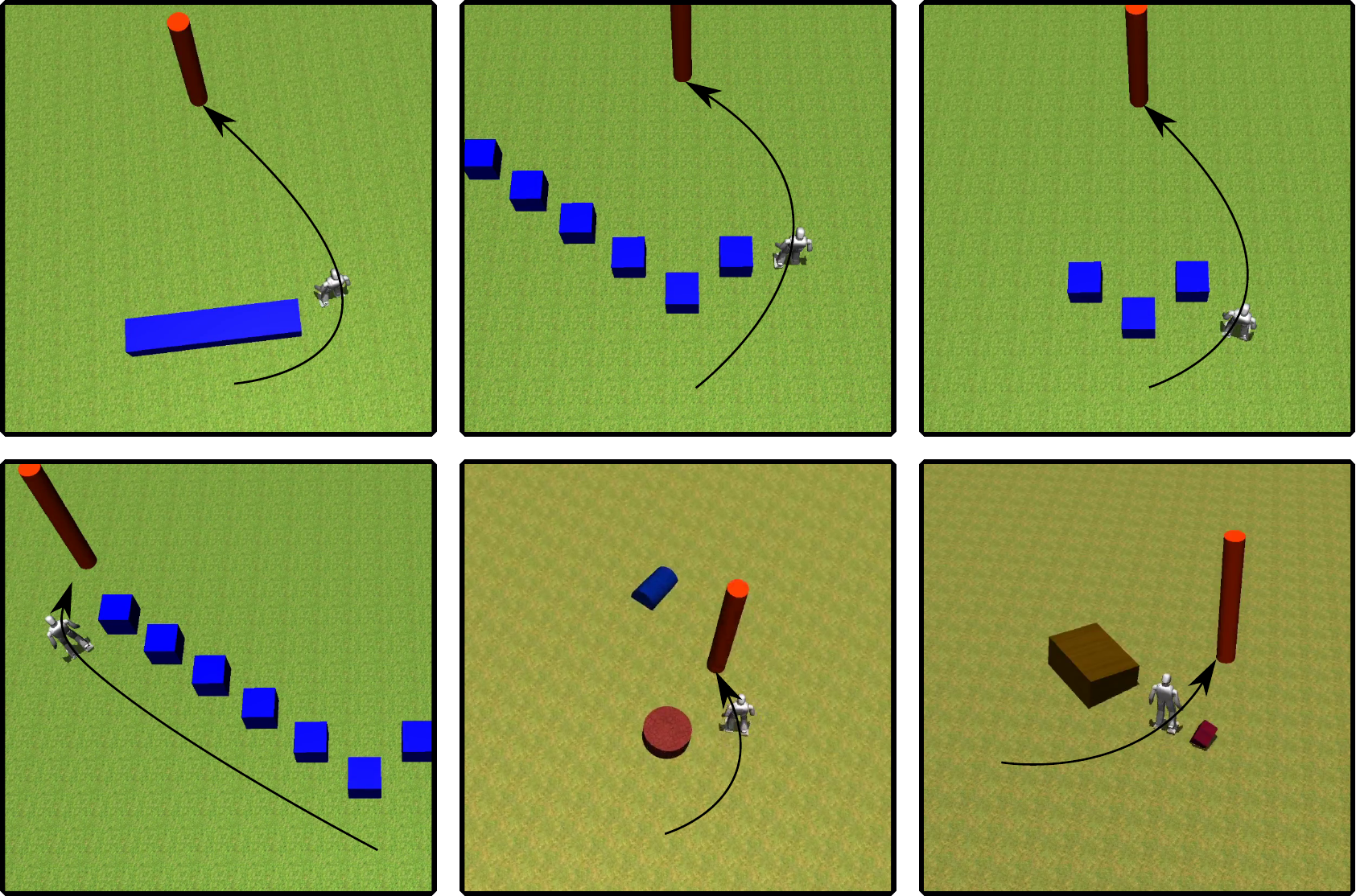}
	\caption{The learned policy successfully navigates in different scenarios where a direct path to the goal pose is blocked by obstacles.
		Observe that the agent is able to circumvent small and large single obstacles.
		Finer control is evidenced in scenarios where the robot is required to go through a passage of obstacles (bottom right)} 
	\label{fig:sim_sequences}
\end{figure}

After training,
the control policy is able to command the robot to reach target poses avoiding obstacles.
Figure~\ref{fig:sim_sequences} shows sample scenarios where the robot navigates collision-free with our learned control policy.
%As anticipated,
The robot circumvents obstacles and goes through narrow passages without falling.
All the environments are presented to the robot for the first time. A model-free DDPG~\cite{ddpg} agent trained for the same amount of steps was not able to solve the environment. 

\begin{figure}[]
	\centering
	\includegraphics[width=0.49\linewidth]{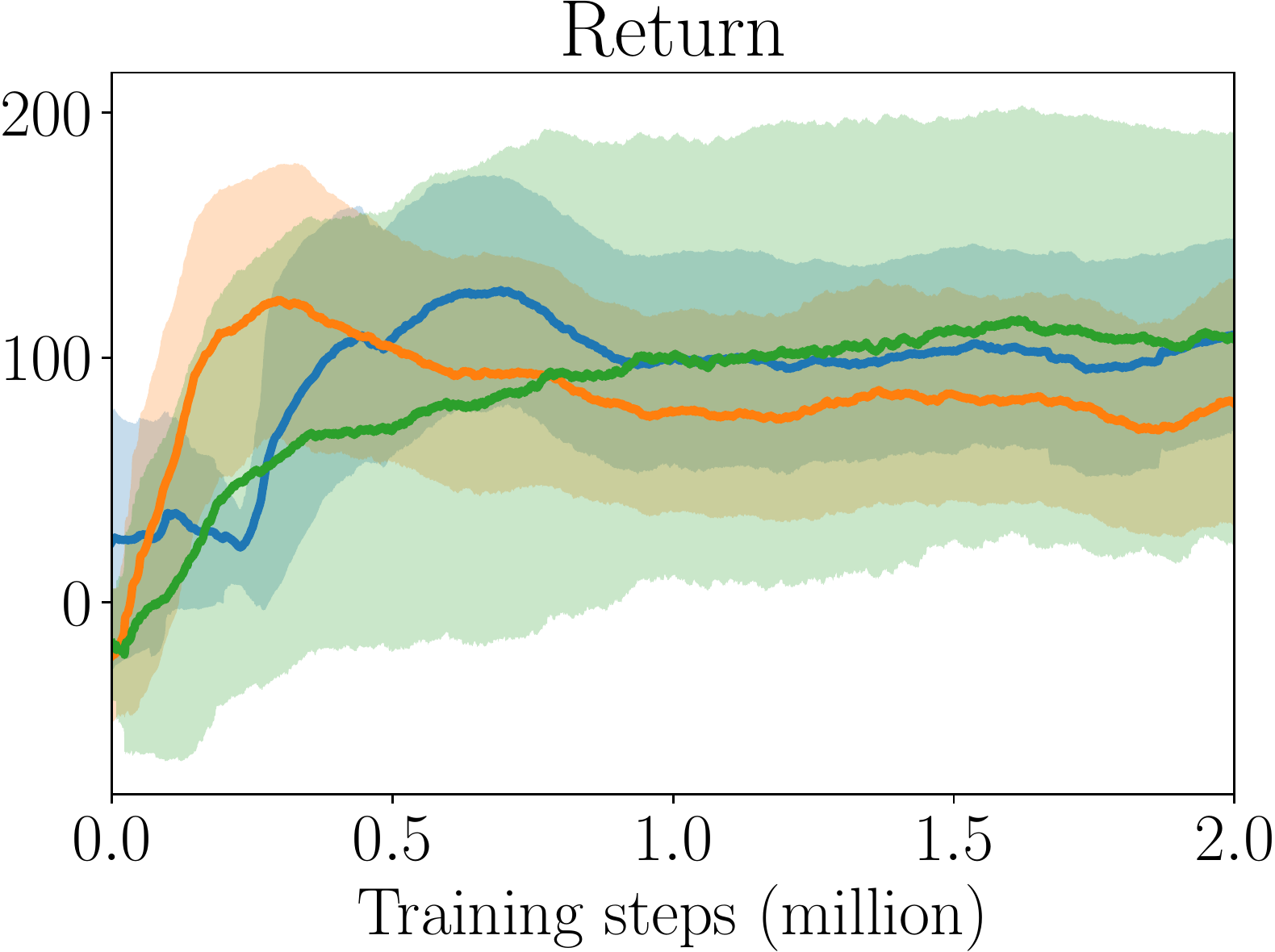}
	\includegraphics[width=0.49\linewidth]{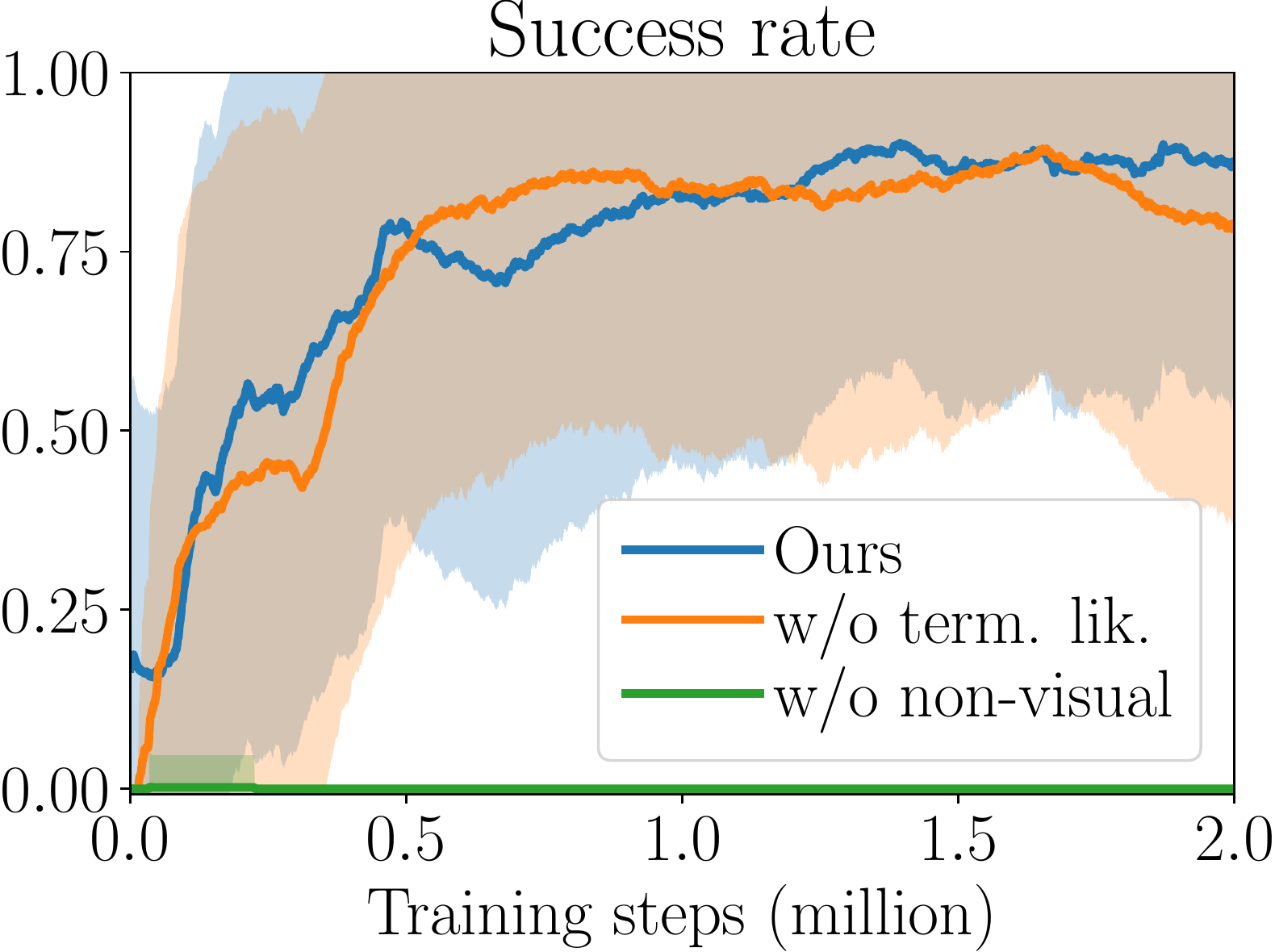}
	\caption{Performance over the number of collected episodes.
		We compare our approach against two ablations by removing the non-visual observations and the termination likelihood predictor. The mean and standard deviation are calculated over the last $250$ collected test episodes.}
	\label{fig:return_curve}
\end{figure}

We compare our approach against two ablated versions of our method.
The first ablation (M\textsubscript{f}) does not include the termination likelihood predictor and the second one (M\textsubscript{nv}) does not consider non-visual observations.
The return and success rate are presented in Fig.~\ref{fig:return_curve}.
Note that although the M\textsubscript{nv} model accumulates more reward compared to our model, 
it is not able to reach the target indicated by its low success rate.
The larger reward is attributed to longer sequences that do not reach the goal
(e.g. the robot might stand in front of the target).
The main contribution in the performance increment is clearly attributed to the introduction of non-visual information,
%which is rather to be expected 
because the agent is not forced to obtain information from the images but it is given directly.
Thus,
the incorporation of non-visual information improves the data-efficiency of the overall approach.
However, 
in complex scenarios where the robot has to back up, 
the agent often only reaches local minima. 
%indicating that the trained policy is mostly reactive.

We evaluate the performance of each of the learned models compared above. 
We generate a set of 100 random scenes sampled as during training. 
The results are presented in Table~\ref{table:results}.
After 2 million steps,
the M\textsubscript{nv} model is not able to successfully complete the task. 
The accumulated reward and the long episodes indicate that the agent does not fall but does not find the target in the given time per episode (\unit[60]{s}).
The higher success rate and reward of our approach compared against the model without termination predictor $\mathcal{F}$ demonstrates the improvement in the sample efficiency by incorporating a predictor for $f_i$.

\begin{table}[]
%	\vspace*{-1ex}
	\centering
	\caption{Success rate, accumulated reward and episode length of our model and ablations over $100$ samples.}
	\vspace{-1ex}
	\label{table:results}
	\begin{tabular}{lcccc}
		\toprule
		 & w/o non-visual & w/o P\textsubscript{term} $f_{i}$ & Ours \\
		\midrule
		Success rate & 0.0 & 0.7 & 0.86 \\
		Acc. reward & 108.50 & 104.37 & 111.59 \\
		Ep. length & 585.36 & 401.15 & 419.77 \\
		\bottomrule
	\end{tabular}
\vspace*{-2ex}
\end{table}

\begin{figure}[]
	\centering
	\includegraphics[width=1.0\linewidth]{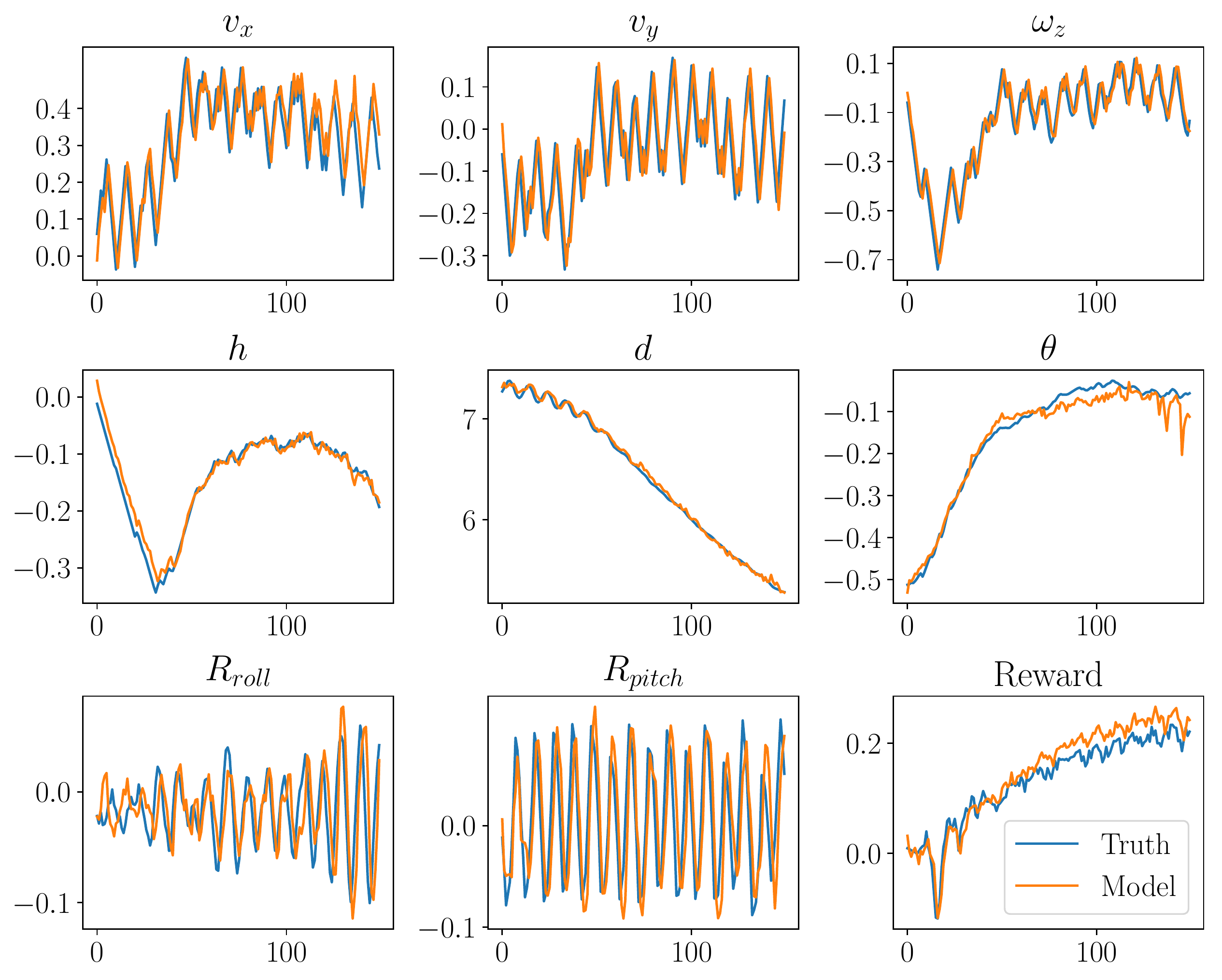}
	\caption{Reconstructed non-visual observations and reward signal for a sampled sequence inferred by our models.
		The better the reconstruction,
		the more realistic are the imagined trajectories used for training.}
	\label{fig:reconstructions}
	\vspace*{-2ex}
\end{figure}

Since the performance of the agent depends on the ability of the model to reproduce the actual observations and reward values from the latent state, 
we evaluate the quality of their reconstruction.
Figure~\ref{fig:reconstructions} shows the reconstructed non-visual observations and reward for a random sequence.
For clarity in the figure, 
only one sequence is shown, 
but other sequences present a similar behavior.
Note how well the model tracks the ground truth signals,
which is to be expected once the models have converged.
The accuracy of the reconstruction is an indicator of the quality of the inferred open-loop trajectories used for training the value and actor networks.

%We compare our approach against Dreamer~\cite{dreamer}, 
%whose model is trained for 3 million steps. 
%On the same computer, 
%the training takes approximately 3 days.
%For the evaluation, 
%we generate a set of 100 random scenes sampled as during training. 

%\subsection{Simulation}

\begin{figure}[b!]
	\centering
	\def\svgwidth{\linewidth}
	\footnotesize
	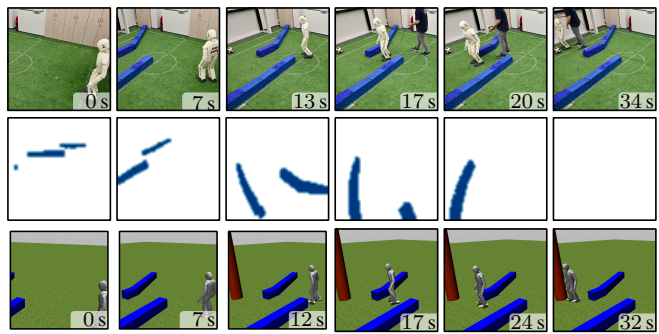
	\caption{The top row shows the real robot navigating a scene by taking a right turn, 
		followed by rotating left to walk through a tight corridor between the obstacles. 
		At the end, 
		the robot walks into the target (soccer ball) by stepping laterally.
		The corresponding segmented image observations of the real experiment are 
		shown in the middle row.
		At the bottom rows,
		a simulated scene is presented,
		where the agent chooses a similar path as the real-world one.}
	\label{fig:hard_sequence}
\end{figure}

\subsection{Real-World Transfer}
The simulated robot is equipped with the same sensors as the real robot.
%Furthermore,
%the visual complexity of the camera images is reduced through semantic segmentation. 
This allows, 
in conjunction with the semantic segmentation,
a real-world transfer with low additional effort and no retraining.
Due to sim-to-real dissimilarities in the robot model, 
joint controllers and contact properties, 
the simulated gait does not behave as the one on the real robot.
To facilitate the sim-to-real transfer,
we inject Gaussian noise,
$\mathcal{N}(0,0.3)$,
on the inferred actions during training.
In addition, 
the gaits are tuned by introducing scaling factors to the inferred actions in order to obtain a similar response in simulation and with the real hardware.
Figure~\ref{fig:hard_sequence} shows a real and a simulated robot performing the same scenario consisting of traversing a narrow passage.
The row in the middle presents the segmented images captured with the real robot,
whereas the bottom row shows the task performed in simulation.
This is a challenging scenario that requires precise actions from the agent to avoid collisions.
The temporal differences between the real and the simulated trajectories are attributed mainly to contact parameters such as frictions.
%Consequently, it is also hard to compare the real-world performance directly to the simulation, since the contact dynamics are not 
%perfectly simulated and the ground-truth data that is necessary for computing the rewards is inaccessible.

\begin{figure}[]
	\centering
	\def\svgwidth{\linewidth}
	\footnotesize
	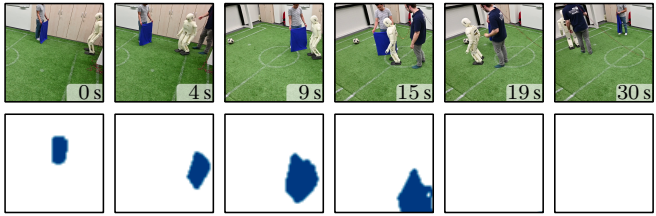
	\caption{The real NimbRo-OP2X robot avoids a moving obstacle that is constantly blocking the path to the target pose (top row).
		The segmented images taken from a first-person view are shown in the bottom row.
	}
	\label{fig:dyn_obstacles}
	\vspace{-2ex}
\end{figure}

Finally, 
the control policy is tested with dynamic obstacles with the real robot.
Figure~\ref{fig:dyn_obstacles} shows snapshots of the robot avoiding a moving obstacle which is blocking the direct path to the target pose.
Note that the policy has not been trained on dynamic obstacles, 
and it has not seen such object shapes.

%% file: figs/compiled/hard_sequence.pdf_tex
%% Creator: Inkscape inkscape 0.92.3, www.inkscape.org
%% PDF/EPS/PS + LaTeX output extension by Johan Engelen, 2010
%% Accompanies image file 'hard_sequence.pdf' (pdf, eps, ps)
%%
%% To include the image in your LaTeX document, write
%%   \input{<filename>.pdf_tex}
%%  instead of
%%   \includegraphics{<filename>.pdf}
%% To scale the image, write
%%   \def\svgwidth{<desired width>}
%%   \input{<filename>.pdf_tex}
%%  instead of
%%   \includegraphics[width=<desired width>]{<filename>.pdf}
%%
%% Images with a different path to the parent latex file can
%% be accessed with the `import' package (which may need to be
%% installed) using
%%   \usepackage{import}
%% in the preamble, and then including the image with
%%   \import{<path to file>}{<filename>.pdf_tex}
%% Alternatively, one can specify
%%   \graphicspath{{<path to file>/}}
%% 
%% For more information, please see info/svg-inkscape on CTAN:
%%   http://tug.ctan.org/tex-archive/info/svg-inkscape
%%
\begingroup%
  \makeatletter%
  \providecommand\color[2][]{%
    \errmessage{(Inkscape) Color is used for the text in Inkscape, but the package 'color.sty' is not loaded}%
    \renewcommand\color[2][]{}%
  }%
  \providecommand\transparent[1]{%
    \errmessage{(Inkscape) Transparency is used (non-zero) for the text in Inkscape, but the package 'transparent.sty' is not loaded}%
    \renewcommand\transparent[1]{}%
  }%
  \providecommand\rotatebox[2]{#2}%
  \newcommand*\fsize{\dimexpr\f@size pt\relax}%
  \newcommand*\lineheight[1]{\fontsize{\fsize}{#1\fsize}\selectfont}%
  \ifx\svgwidth\undefined%
    \setlength{\unitlength}{1184.96510836bp}%
    \ifx\svgscale\undefined%
      \relax%
    \else%
      \setlength{\unitlength}{\unitlength * \real{\svgscale}}%
    \fi%
  \else%
    \setlength{\unitlength}{\svgwidth}%
  \fi%
  \global\let\svgwidth\undefined%
  \global\let\svgscale\undefined%
  \makeatother%
  \begin{picture}(1,0.5006884)%
    \lineheight{1}%
    \setlength\tabcolsep{0pt}%
    \put(-0.10875618,0.56712268){\color[rgb]{0,0,0}\makebox(0,0)[lt]{\begin{minipage}{1.20166282\unitlength}\raggedright \end{minipage}}}%
    \put(0,0){\includegraphics[width=\unitlength,page=1]{hard_sequence.pdf}}%
    \put(0.15637597,0.34569886){\color[rgb]{0,0,0}\makebox(0,0)[rt]{\lineheight{1.25}\smash{\begin{tabular}[t]{r}$0\,$s\end{tabular}}}}%
    \put(0,0){\includegraphics[width=\unitlength,page=2]{hard_sequence.pdf}}%
    \put(0.32066315,0.34195507){\color[rgb]{0,0,0}\makebox(0,0)[rt]{\lineheight{1.25}\smash{\begin{tabular}[t]{r}$7\,$s\end{tabular}}}}%
    \put(0.4932022,0.34569886){\color[rgb]{0,0,0}\makebox(0,0)[rt]{\lineheight{1.25}\smash{\begin{tabular}[t]{r}$13\,$s\end{tabular}}}}%
    \put(0.66023927,0.34569886){\color[rgb]{0,0,0}\makebox(0,0)[rt]{\lineheight{1.25}\smash{\begin{tabular}[t]{r}$17\,$s\end{tabular}}}}%
    \put(0.82819382,0.34569886){\color[rgb]{0,0,0}\makebox(0,0)[rt]{\lineheight{1.25}\smash{\begin{tabular}[t]{r}$20\,$s\end{tabular}}}}%
    \put(0.99614829,0.34569886){\color[rgb]{0,0,0}\makebox(0,0)[rt]{\lineheight{1.25}\smash{\begin{tabular}[t]{r}$34\,$s\end{tabular}}}}%
    \put(0.99614836,0.00523816){\color[rgb]{0,0,0}\makebox(0,0)[rt]{\lineheight{1.25}\smash{\begin{tabular}[t]{r}$32\,$s\end{tabular}}}}%
    \put(0.82819382,0.00523823){\color[rgb]{0,0,0}\makebox(0,0)[rt]{\lineheight{1.25}\smash{\begin{tabular}[t]{r}$24\,$s\end{tabular}}}}%
    \put(0.66023934,0.00523823){\color[rgb]{0,0,0}\makebox(0,0)[rt]{\lineheight{1.25}\smash{\begin{tabular}[t]{r}$17\,$s\end{tabular}}}}%
    \put(0.48865332,0.00880375){\color[rgb]{0,0,0}\makebox(0,0)[rt]{\lineheight{1.25}\smash{\begin{tabular}[t]{r}$12\,$s\end{tabular}}}}%
    \put(0.32069892,0.0088035){\color[rgb]{0,0,0}\makebox(0,0)[rt]{\lineheight{1.25}\smash{\begin{tabular}[t]{r}$7\,$s\end{tabular}}}}%
    \put(0.15274443,0.00880379){\color[rgb]{0,0,0}\makebox(0,0)[rt]{\lineheight{1.25}\smash{\begin{tabular}[t]{r}$0\,$s\end{tabular}}}}%
  \end{picture}%
\endgroup%

%% file: figs/compiled/real_dyn_sequence.pdf_tex
%% Creator: Inkscape inkscape 0.92.3, www.inkscape.org
%% PDF/EPS/PS + LaTeX output extension by Johan Engelen, 2010
%% Accompanies image file 'real_dyn_sequence.pdf' (pdf, eps, ps)
%%
%% To include the image in your LaTeX document, write
%%   \input{<filename>.pdf_tex}
%%  instead of
%%   \includegraphics{<filename>.pdf}
%% To scale the image, write
%%   \def\svgwidth{<desired width>}
%%   \input{<filename>.pdf_tex}
%%  instead of
%%   \includegraphics[width=<desired width>]{<filename>.pdf}
%%
%% Images with a different path to the parent latex file can
%% be accessed with the `import' package (which may need to be
%% installed) using
%%   \usepackage{import}
%% in the preamble, and then including the image with
%%   \import{<path to file>}{<filename>.pdf_tex}
%% Alternatively, one can specify
%%   \graphicspath{{<path to file>/}}
%% 
%% For more information, please see info/svg-inkscape on CTAN:
%%   http://tug.ctan.org/tex-archive/info/svg-inkscape
%%
\begingroup%
  \makeatletter%
  \providecommand\color[2][]{%
    \errmessage{(Inkscape) Color is used for the text in Inkscape, but the package 'color.sty' is not loaded}%
    \renewcommand\color[2][]{}%
  }%
  \providecommand\transparent[1]{%
    \errmessage{(Inkscape) Transparency is used (non-zero) for the text in Inkscape, but the package 'transparent.sty' is not loaded}%
    \renewcommand\transparent[1]{}%
  }%
  \providecommand\rotatebox[2]{#2}%
  \newcommand*\fsize{\dimexpr\f@size pt\relax}%
  \newcommand*\lineheight[1]{\fontsize{\fsize}{#1\fsize}\selectfont}%
  \ifx\svgwidth\undefined%
    \setlength{\unitlength}{1175.92390436bp}%
    \ifx\svgscale\undefined%
      \relax%
    \else%
      \setlength{\unitlength}{\unitlength * \real{\svgscale}}%
    \fi%
  \else%
    \setlength{\unitlength}{\svgwidth}%
  \fi%
  \global\let\svgwidth\undefined%
  \global\let\svgscale\undefined%
  \makeatother%
  \begin{picture}(1,0.32426532)%
    \lineheight{1}%
    \setlength\tabcolsep{0pt}%
    \put(0,0){\includegraphics[width=\unitlength,page=1]{real_dyn_sequence.pdf}}%
    \put(0.15007456,0.17552133){\color[rgb]{0,0,0}\makebox(0,0)[rt]{\lineheight{1.25}\smash{\begin{tabular}[t]{r}$0\,$s\end{tabular}}}}%
    \put(0,0){\includegraphics[width=\unitlength,page=2]{real_dyn_sequence.pdf}}%
    \put(0.31932035,0.17552133){\color[rgb]{0,0,0}\makebox(0,0)[rt]{\lineheight{1.25}\smash{\begin{tabular}[t]{r}$4\,$s\end{tabular}}}}%
    \put(0,0){\includegraphics[width=\unitlength,page=3]{real_dyn_sequence.pdf}}%
    \put(0.48856612,0.17552133){\color[rgb]{0,0,0}\makebox(0,0)[rt]{\lineheight{1.25}\smash{\begin{tabular}[t]{r}$9\,$s\end{tabular}}}}%
    \put(0,0){\includegraphics[width=\unitlength,page=4]{real_dyn_sequence.pdf}}%
    \put(0.65781193,0.17552133){\color[rgb]{0,0,0}\makebox(0,0)[rt]{\lineheight{1.25}\smash{\begin{tabular}[t]{r}$15\,$s\end{tabular}}}}%
    \put(0,0){\includegraphics[width=\unitlength,page=5]{real_dyn_sequence.pdf}}%
    \put(0.82705773,0.17552133){\color[rgb]{0,0,0}\makebox(0,0)[rt]{\lineheight{1.25}\smash{\begin{tabular}[t]{r}$19\,$s\end{tabular}}}}%
    \put(0,0){\includegraphics[width=\unitlength,page=6]{real_dyn_sequence.pdf}}%
    \put(0.99630354,0.17552133){\color[rgb]{0,0,0}\makebox(0,0)[rt]{\lineheight{1.25}\smash{\begin{tabular}[t]{r}$30\,$s\end{tabular}}}}%
    \put(0,0){\includegraphics[width=\unitlength,page=7]{real_dyn_sequence.pdf}}%
  \end{picture}%
\endgroup%

%% file: conclusion.tex
\section{Conclusion}
\label{sec:conclusion}
In this paper, 
we have proposed a novel approach for learning mapless navigation around obstacles based on visual and non-visual observations.
We have demonstrated that the incorporation of a termination likelihood predictor increases the data-efficiency of the approach.
%Our model exceeds the performance of the Dreamer approach in sample-efficiency and mean reward.
In addition, 
we have shown that our model produces a robust policy that can be successfully transferred to a real humanoid robot.

In the future, 
we would like to extend our approach to incorporate hierarchies.
Multiple consistent policies are envisioned to solve more complex tasks that require long-term planning.
Additionally, 
learning local and global maps seems to be a promising alternative to provide the agent with more sophisticated navigation skills,
such as remembering dead ends.
More dynamic scenarios where multiple objects move simultaneously require the agent to track and estimate the velocities of the moving bodies,
which also states an interesting problem to enrich our approach.